\documentclass[conference,11pt]{IEEEtran}
\usepackage{cite}
\usepackage{amsmath,amssymb,amsfonts}
\usepackage{algorithmic}
\usepackage{graphicx}
\usepackage{textcomp}
\usepackage{xcolor}
\usepackage{booktabs}
\usepackage{stfloats}
\usepackage{threeparttable}
\usepackage{url}
\usepackage{xurl}
\usepackage{pifont}
\newcommand{\cmark}{\ding{51}}
\newcommand{\xmark}{\ding{55}}
\usepackage{makecell}
\usepackage[top=0.75in,bottom=1.15in,left=0.64in,right=0.64in]{geometry}

\begin{document}
\setlength{\columnsep}{0.25in}
\fontsize{11pt}{13.2pt}\selectfont

\title{Auto-Unrolled Proximal Gradient Descent: An AutoML Approach to Interpretable Waveform Optimization}

\author{\IEEEauthorblockN{Ahmet Kaplan, \textit{Member, IEEE}}
\IEEEauthorblockA{Department of Computer Science\\
Istanbul Medipol University\\
Istanbul, Türkiye \\
ahmet.kaplan@medipol.edu.tr}
}

\maketitle

\begin{abstract}
Deep-unfolded networks for wireless beamforming require manual selection of unrolling depth and step-size schedules, limiting adaptability without retraining. We propose Auto-Unrolled PGD (Auto-PGD), a physics-structured beamforming framework that embeds the proximal gradient descent (PGD) algorithm into an $L$-layer neural network, where each layer performs one learnable PGD step with a power-feasible projection, and uses AutoGluon's Tree-structured Parzen Estimator (TPE) to autonomously discover the optimal depth and per-layer step-size schedule. Evaluated over 5 random seeds on a multiple-input single-output (MISO) downlink ($M{=}8$, $K{=}4$, Rayleigh fading), Auto-PGD achieves $98.0\%$ and $98.8\%$ of a 200-iteration classical PGD solver with only $N{=}100$ and $N{=}1000$ training samples ($14.52{\pm}0.12$ and $14.64{\pm}0.00$ bits/s/Hz), using 5 unrolled layers and a single forward pass, a $40{\times}$ reduction in gradient evaluations relative to classical PGD at inference time. Auto-PGD outperforms both a black-box multilayer perceptron (MLP) ($10.7\%$ of PGD at $N{=}100$) and fixed-depth PGD-Net ($85.8\%$) by large margins, with near-zero variance at $N{=}1000$. 
\end{abstract}


\section{Introduction}
Real-time optimization in signal processing often faces a trade-off between the mathematical rigor of iterative solvers and the computational speed of deep learning. Sixth-generation (6G) wireless networks target peak data rates of 1 Tbps and, more critically, ultra-reliable low-latency communication (URLLC) with end-to-end delays below 0.1 ms. Achieving such stringent requirements necessitates the real-time optimization of complex waveforms and beamforming vectors in highly dynamic environments.

Traditionally, this optimization relies on iterative solvers such as Proximal Gradient Descent (PGD) or Weighted Minimum Mean Square Error (WMMSE) \cite{shi2011wmmse}. While mathematically rigorous and interpretable, these algorithms often require hundreds of iterations to converge, making them computationally prohibitive for the microsecond-scale scheduling intervals of 6G. Conversely, standard Deep Learning (DL) architectures offer near-instantaneous inference but operate as "black boxes." These models typically demand massive, diverse datasets to learn physical propagation laws and lack the formal guarantees required for safety-critical URLLC applications in 6G.

Despite the promise of deep unfolding, three open challenges limit its practical deployment. First, the unrolling depth $L$ is fixed manually and held constant across all data regimes, yet the optimal $L$ is inherently data-quantity-dependent: shallow networks generalize better with scarce data, while deeper networks benefit from abundant data. Second, the per-layer step-size schedule is either hand-tuned or initialized uniformly, with no systematic search over the joint configuration space of depth and step sizes. Third, no existing work has applied automated hyperparameter search to physics-structured unrolled networks for physical-layer beamforming, leaving the potential of joint depth--step-size optimization unexplored.

To address these challenges, we propose Auto-Unrolled PGD (Auto-PGD). By unrolling the iterations of a PGD solver into a structured neural network, we embed physical domain knowledge directly into the architecture. The optimal unrolling depth is data-quantity-dependent---5 layers suffice at $N{\leq}1000$ while deeper networks are needed at larger $N$---a design choice that prior work fixes by convention rather than by data-driven optimization. We integrate AutoGluon, an AutoML engine that jointly searches over depth and step-size initialization via Tree-structured Parzen Estimation (TPE), to eliminate this manual choice and exploit the data-dependent optimum automatically.

This paper makes three contributions. First, we formulate an unrolled PGD network for 6G MISO beamforming with a power-feasible proximal projection that is structurally incapable of violating the transmit power constraint by design. Second, we show that AutoGluon's TPE-based hyperparameter optimization (HPO) autonomously discovers that only 5 unrolled layers suffice in the data-scarce regime ($N{\leq}1000$), achieving $14.3\%$--$18.2\%$ higher sum-rate than a fixed-depth PGD-Net (10 layers, no HPO) that uses the same physics structure. Third, we show that Auto-PGD achieves $98.0\%$ and $98.8\%$ of classical PGD spectral efficiency at $N{=}100$ and $N{=}1000$ (5-seed means $14.52{\pm}0.12$ and $14.64{\pm}0.00$ bits/s/Hz), outperforming both black-box MLP ($10.7\%$) and fixed-depth PGD-Net ($85.8\%$), with a $40{\times}$ reduction in per-channel gradient evaluations at inference time relative to classical PGD.

The remainder of this paper is organized as follows. Section~II reviews related work. Section~III provides background on PGD and deep unfolding. Section~IV formulates the beamforming problem. Section~V describes the proposed Auto-PGD methodology. Section~VI details the experimental design. Section~VII presents results. Section~VIII discusses interpretability. Section~IX concludes.

\section{Related Work}

\textbf{Deep Unfolding for Wireless Beamforming.}
Deep unfolding (DU) bridges iterative optimization and deep learning by truncating an iterative algorithm to $K$ layers and treating its parameters as learnable \cite{balatsoukas2019deepunfolding}. Early work unrolled the Iterative Soft Thresholding Algorithm (LISTA) for sparse recovery \cite{gregor2010lista}; DU was subsequently extended to multiple-input multiple-output (MIMO) precoding via the Iterative Algorithm Induced Deep-Unfolding Neural Network (IAIDNN), which unrolls WMMSE into a trainable layer sequence for multiuser MIMO systems \cite{hu2020deepunfolding}. Deep-unfolded WMMSE variants for downlink beamforming further replace fixed WMMSE iterations with learned weight matrices, achieving faster convergence with fewer layers \cite{pellaco2021deep}. More recently, DU has been applied to 6G-oriented scenarios including stacked intelligent metasurface (SIM)-assisted MISO, robust fractional-programming throughput maximization, and hybrid learned-optimization strategies \cite{ibrahim2026deepunfolding,bui2026deepunfolded,zhu2026deepfp}; Shlezinger et al. provide a comprehensive survey of DU theory and design guidelines \cite{shlezinger2025deepunfolding}. A common limitation across all these works is that the unrolling depth $L$ is fixed manually as a design constant, with no principled data-driven selection mechanism.

\textbf{AutoML and Hyperparameter Optimization for Physical-Layer Systems.}
Tree-structured Parzen Estimation (TPE), the HPO algorithm we employ, was introduced by Bergstra et al.~\cite{bergstra2011tpe} and models the prior over configurations as two density estimators, guiding search toward high-performing regions more efficiently than random or grid search. Automated Machine Learning (AutoML) frameworks such as AutoGluon \cite{erickson2020autogluon} build on TPE and related methods to automate model selection and hyperparameter tuning across domains. In the wireless domain, AutoML has been applied to network configuration optimization \cite{filippou2023automl} and neural architecture search for communication systems \cite{wang2024nas}. However, no prior work has applied AutoML to jointly optimize the unrolling depth and per-layer step-size schedule of a physics-structured DU network for physical-layer beamforming. This gap motivates the proposed Auto-PGD framework, which treats $L$ and the step-size initialization as HPO variables rather than fixed design constants.

\section{Background}
\subsection{Proximal Gradient Descent and Deep Unfolding}
PGD solves
\begin{equation}
  \min_x \; f(x) + g(x)
\end{equation}
via the iteration
\begin{equation}
  x^{(k+1)} = \operatorname{prox}_{\eta^{(k)}g}\!\left(x^{(k)} - \eta^{(k)}\nabla f\!\left(x^{(k)}\right)\right),
\end{equation}
where $f$ is differentiable and $g$ encodes constraints. Deep unfolding truncates this recursion to $L$ layers and treats $\{\eta^{(\ell)}\}_{\ell=1}^{L}$ as learnable weights \cite{balatsoukas2019deepunfolding}. The resulting hypothesis space
\begin{equation}
  \mathcal{H}_{\mathrm{DU}} = \left\{\operatorname{prox}_{\eta g}\!\circ(I-\eta\nabla f)^L \;\middle|\; \eta\in\mathbb{R}^+\right\}
\end{equation}
is far smaller than that of a general deep neural network (DNN), requiring significantly fewer training samples to generalize \cite{shlezinger2025deepunfolding}. The optimal $L$ is data-quantity-dependent and is determined by the AutoGluon HPO search described in Section~V.

\subsection{Weighted Minimum Mean Square Error (WMMSE)}
WMMSE \cite{shi2011wmmse} is the canonical iterative baseline for sum-rate maximization in MISO/MIMO systems. It alternates between updating auxiliary weights and solving a sequence of weighted least-squares subproblems, converging to a stationary point of the sum-rate objective. WMMSE requires tens to hundreds of per-channel iterations and is computationally prohibitive at 6G latency constraints. In our experiments, WMMSE (100 iterations) serves as an independent oracle achieving 14.82 bits/s/Hz, which closely matches Classical PGD (200 iterations, 14.81 bits/s/Hz), confirming that both solvers have converged to the near-optimal solution and validating their use as upper-bound references.

\subsection{LISTA and Structured Deep Unfolding Baselines}
The Learned ISTA (LISTA) \cite{gregor2010lista} is the seminal deep unfolding architecture, which unrolls the Iterative Shrinkage-Thresholding Algorithm (ISTA) for sparse signal recovery by replacing the fixed shrinkage threshold with a learned parameter per layer. LISTA demonstrates that unrolling an iterative algorithm into a truncated network can dramatically accelerate convergence with far fewer layers than the original algorithm requires. In our setting, LISTA is adapted to the sum-rate maximization objective as a structured fixed-depth baseline: it shares the unrolling philosophy with Auto-PGD but lacks the power-feasibility projection and the AutoML-driven depth selection, making it a direct ablation point for the value of physics-structured constraints.

\section{Problem Formulation}

We consider a 6G downlink Multi-Input Single-Output (MISO) system where a Base Station (BS) equipped with $M$ antennas serves $K$ single-antenna users. The signal received by the $k$-th user is given by:
\begin{equation}
y_k = \mathbf{h}_k^H \mathbf{w}_k s_k + \sum_{j \neq k} \mathbf{h}_k^H \mathbf{w}_j s_j + n_k
\end{equation}
where $s_k$ is the transmitted symbol for user $k$, $n_k$ is the additive white Gaussian noise (AWGN; $n_k \sim \mathcal{CN}(0, \sigma^2)$), $\mathbf{h}_k \in \mathbb{C}^{M}$ is the channel vector, and $\mathbf{w}_k \in \mathbb{C}^{M}$ is the beamforming vector.

\subsection{Optimization Objective}
The primary goal is to maximize the system Sum-Rate under a total power constraint. The optimization problem is formulated as:
\begin{align}
\max_{\mathbf{W}} \quad & \sum_{k=1}^K \log_2 \left( 1 + \frac{|\mathbf{h}_k^H \mathbf{w}_k|^2}{\sum_{j \neq k} |\mathbf{h}_k^H \mathbf{w}_j|^2 + \sigma^2} \right) \\
\text{s.t.} \quad & \sum_{k=1}^K \|\mathbf{w}_k\|^2 \leq P_{max}
\end{align}
where $\mathbf{W} = [\mathbf{w}_1, \dots, \mathbf{w}_K]$ is the beamforming matrix and $P_{max}$ is the maximum transmit power.

We decompose $\Phi(\mathbf{W}) = f(\mathbf{W}) + g(\mathbf{W})$, where $f$ is the negative sum-rate (differentiable) and $g$ is the power-constraint indicator ($g(\mathbf{W})=0$ if $\sum\|\mathbf{w}_k\|^2\leq P_{\max}$, else $+\infty$). The proximal operator of $g$ is a Frobenius-ball projection:
\begin{equation}
\text{prox}_{\eta g}(\mathbf{Z}) = \frac{\sqrt{P_{\max}}}{\max(\sqrt{P_{\max}}, \|\mathbf{Z}\|_F)} \mathbf{Z}
\end{equation}
guaranteeing power-feasible outputs by construction. This decomposition renders the problem PGD-compatible and motivates the unrolled architecture described in Section~V.

\section{Proposed Methodology}
\subsection{Deep Unrolling of PGD}
We unroll $L$ iterations of PGD into an $L$-layer feed-forward network, where $L$ is determined by the AutoGluon search described below. Each layer applies a scalar learned step-size $\eta^{(\ell)}$ to the sum-rate gradient:
\begin{equation}
\mathbf{W}^{(\ell+1)} = \text{prox}_{\eta^{(\ell)} g}\!\left(\mathbf{W}^{(\ell)} - \eta^{(\ell)} \nabla f(\mathbf{W}^{(\ell)})\right)
\end{equation}
where $\text{prox}_{\eta g}$ is the Frobenius-ball projection defined in Section~IV. The network is initialized with a Zero-Forcing (ZF) beamforming matrix, and the learnable parameters $\theta = \{\eta^{(\ell)}\}_{\ell=1}^{L}$ number $O(L)$---orders of magnitude fewer than a comparable MLP ($O(10^5)$). The HPO search also evaluates a \emph{hybrid layer} variant that adds a learnable gradient-transformation matrix; the standard PGD layer (scalar step-size only) was selected at all tested sizes, confirming that the physics structure suffices without additional parameters.

\subsection{AutoML Integration via AutoGluon}
To automatically determine $L$ and the per-layer step-size initialization, we use AutoGluon with an Optuna Tree-structured Parzen Estimator (TPE) sampler. The key insight is that the optimal $L$ is data-quantity-dependent---5 layers suffice at $N{\leq}1000$ while deeper networks are beneficial at larger $N$---a choice that manual design fixes at a single value. TPE exploits the history of evaluated configurations to guide search more efficiently than random or grid search.

The HPO objective is $\max_{L,\,\eta^{(0)},\,\text{opt},\,\text{sched}} r_{\text{val}}$ subject to inference latency $\tau < 0.1$\,ms, where $r_{\text{val}}$ is the sum-rate on a held-out 10\% validation split of the training data.

\begin{figure}[htbp]
    \centering
    \includegraphics[width=0.48\textwidth]{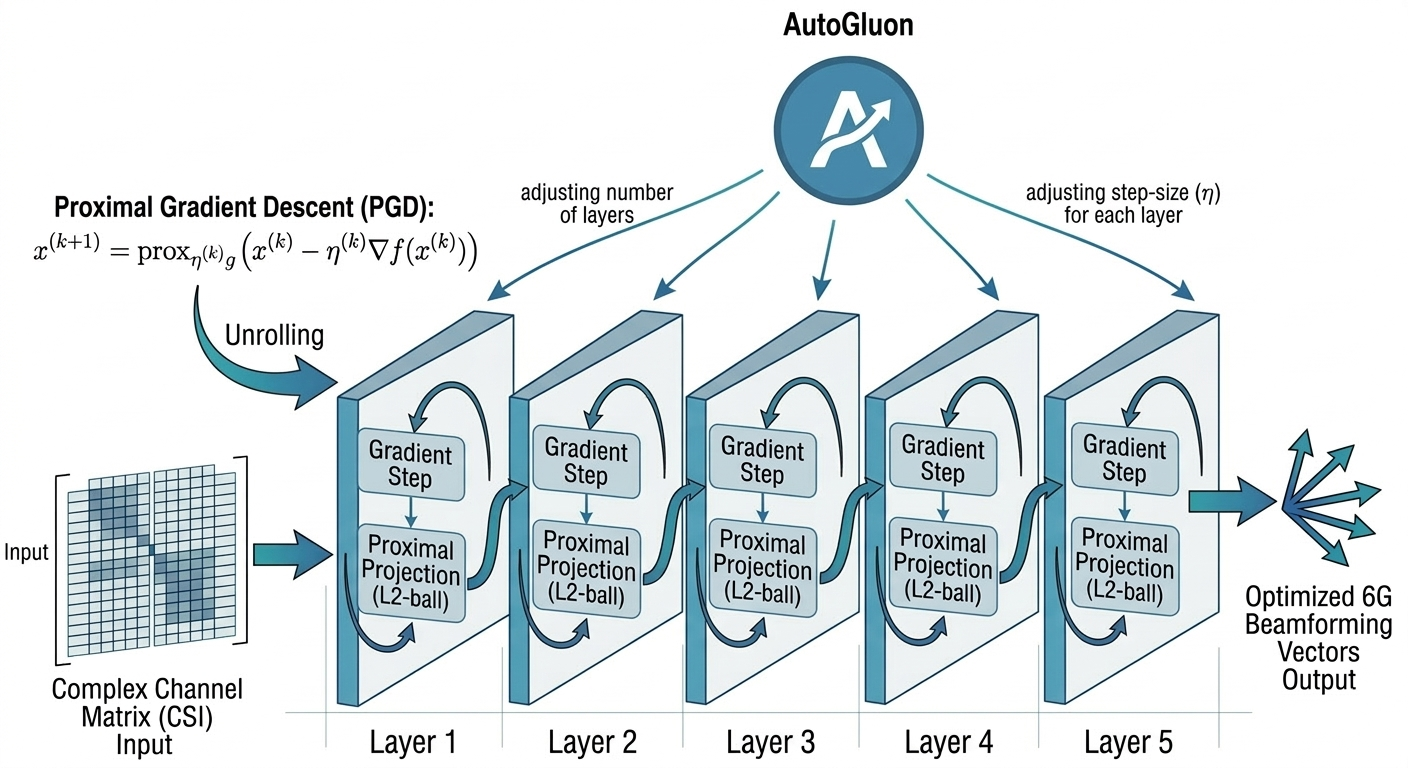}
    \caption{Abstract workflow of Auto-Unrolled Proximal Gradient Descent: classical PGD is transformed into a deep-unfolded network, AutoGluon automates architecture search and hyperparameter optimization, and the system outputs optimized beamforming vectors for wireless waveform optimization.}
    \label{fig:autogluon_diagram}
\end{figure}

Figure~\ref{fig:autogluon_diagram} illustrates the abstract workflow of the proposed Auto-Unrolled Proximal Gradient Descent framework. The diagram highlights the transformation of classical PGD into a deep-unfolded network, the integration of AutoGluon for automated architecture search and hyperparameter optimization, and the generation of optimized beamforming vectors for wireless waveform optimization. 

\section{Experimental Design}

\subsection{Simulation Environment}
We simulate a downlink MISO system with $M=8$ antennas and $K=4$ users. Channels follow $\mathbf{h}_k\sim\mathcal{CN}(0,\mathbf{I})$ (Rayleigh fading, 3GPP UMi, 28\,GHz). AWGN power $\sigma^2=0.1$. All key simulation and training parameters are summarized in Table~\ref{tab:sim_params}.

\begin{table}[htbp]
\caption{Simulation and Training Parameters}
\label{tab:sim_params}
\begin{center}
\begin{tabular}{l l l}
\toprule
Parameter & Symbol & Value \\
\midrule
Number of BS antennas & $M$ & 8 \\
Number of users & $K$ & 4 \\
Maximum transmit power & $P_{\max}$ & 1 \\
AWGN variance & $\sigma^2$ & 0.1 \\
Channel model & --- & Rayleigh, i.i.d.\ $\mathcal{CN}(0,\mathbf{I})$ \\
Carrier frequency & --- & 28\,GHz (3GPP UMi) \\
Test set size & $N_{\text{test}}$ & 5{,}000 \\
Random seeds & --- & $\{42,\,678,\,888,\,123,\,456\}$ \\
HPO trials ($N{=}10^2$, $10^3$) & --- & 50 \\
HPO trials ($N{=}10^4$) & --- & 20 (truncated) \\
HPO trials ($N{=}5\!\times\!10^4$) & --- & 5 (truncated) \\
HPO fixed seed & --- & 0 \\
Classical PGD iterations & --- & 200 \\
WMMSE iterations & --- & 100 \\
\bottomrule
\end{tabular}
\end{center}
\end{table}

\subsection{Dataset and Baselines}
Training sizes $n\in\{10^2,10^3,10^4,5\!\times\!10^4\}$ are drawn from a master set of 50,000 Rayleigh samples; inputs are $[\text{Re}(\mathbf{H}),\text{Im}(\mathbf{H})]$ and labels are ZF-initialized beamforming matrices. Baselines: Classical PGD (200 iterations, per-channel), WMMSE \cite{shi2011wmmse} (100 iterations, per-channel; included as an independent oracle to validate that PGD-200 has converged to the near-optimal solution), MLP (5-layer, 256 hidden), LISTA (unrolled soft-thresholding adapted to sum-rate maximization as a structured fixed-depth baseline \cite{gregor2010lista}), PGD-Net (unrolled PGD, 10 layers, no HPO), and ZF (closed-form).

\subsection{AutoML Search via AutoGluon}
\label{sec:hpo}
The HPO search space covers: network depth $K_{\text{unroll}}\in[3,25]$, initial step-size $\eta^{(0)}$ log-uniform in $[10^{-4},10^{-1}]$, optimizer $\in\{\texttt{adam},\texttt{sgd}\}$, and scheduler $\in\{\texttt{cosine},\texttt{step}\}$. The objective is to maximize validation sum-rate subject to $\tau<0.1$\,ms. HPO uses Optuna's TPE sampler with a random-search fallback. The configured budget is 50 trials per size; in practice, $N\in\{10^2,10^3\}$ completed all 50 trials, while $N{=}10^4$ completed 20 trials and $N{=}5\!\times\!10^4$ completed 5 trials due to the higher per-trial training cost. Results at these two larger sizes should therefore be interpreted as lower bounds on Auto-PGD's achievable performance under full HPO.

\subsection{Reproducibility and Statistical Analysis}
All experiments are fully reproducible. Code is at \url{https://github.com/Ahmet-Kaplan/autogluon_waveform}. Results are averaged over seeds $\{42, 678, 888, 123, 456\}$ (5 seeds). HPO uses a fixed seed of 0. Training uses a sum-loss $\mathcal{L}=-\sum_i r_i$ rather than mean-loss to keep gradients batch-size invariant.

\textbf{Statistical analysis.} With $n{=}5$ seeds, we report 95\% confidence intervals (CIs) using the $t$-distribution ($t_{0.025,\,\mathrm{df}{=}4} = 2.776$), which is valid for any continuous distribution at this sample size and does not require a normality assumption. The 95\% CI for Auto-PGD at $N{=}100$ is $[14.37,\,14.67]$ bits/s/Hz, and for PGD-Net is $[11.59,\,13.81]$ bits/s/Hz; at $N{=}1000$, the Auto-PGD CI is $[14.63,\,14.64]$ (near-zero width) versus PGD-Net's $[11.57,\,13.21]$. In both cases the CIs are non-overlapping, providing statistical evidence that the observed performance gap is not attributable to seed-to-seed variance.

\section{Experimental Results}

\subsection{Quantitative Performance Comparison}
Table~\ref{tab:sum_rate} reports the test sum-rate (bits/s/Hz) achieved by all six methods across four training-set sizes on a fixed test set of $N_{\text{test}}=5000$ Rayleigh fading channel realizations ($M=8$, $K=4$, $P_{\max}=1$, $\sigma^2=0.1$). The Zero-Forcing and Classical PGD results are independent of training size. WMMSE (100 iterations) achieves $14.82$ bits/s/Hz—effectively matching Classical PGD—and is included to confirm that the 200-iteration PGD reference is representative of the oracle optimum.

\begin{table*}[b]
\caption{Test Sum-Rate (bits/s/Hz) vs.\ Training Set Size.}
\label{tab:sum_rate}
\begin{center}
\begin{threeparttable}
\begin{tabular}{c c c c c c c c}
\toprule
$N$ & ZF & PGD (200-it) & WMMSE (100-it) & MLP & LISTA & PGD-Net & Auto-PGD \\
\midrule
$10^2$        & 13.33 & 14.81 & 14.82 & $1.58{\pm}0.03$ & $2.26{\pm}0.37$ & $12.70{\pm}0.89$ & $\mathbf{14.52}{\pm}0.12$ \\
$10^3$        & 13.33 & 14.81 & 14.82 & $2.38{\pm}0.14$ & $6.52{\pm}0.02$ & $12.39{\pm}0.66$ & $\mathbf{14.64}{\pm}0.00$ \\
$10^4$        & 13.33 & 14.81 & 14.82 & $4.47{\pm}0.04$ & $6.15{\pm}0.01$ & $11.80{\pm}0.09$ & $\mathbf{13.58}{\pm}0.15$ \\
$5\times10^4$ & 13.33 & 14.81 & 14.82 & $6.45{\pm}0.29$ & $6.98{\pm}0.01$ & $11.65{\pm}0.09$ & $\mathbf{11.89}{\pm}0.06$ \\
\bottomrule
\end{tabular}
\begin{tablenotes}
\small
\item Mean $\pm$ std over 5 random seeds $\{42,678,888,123,456\}$ (all training sizes). ZF, Classical PGD, and WMMSE are training-size independent. WMMSE \cite{shi2011wmmse} (100 iterations) achieves 14.82 bits/s/Hz. System: $M{=}8$, $K{=}4$, $P_{\max}{=}1$, $\sigma^2{=}0.1$, $N_{\text{test}}{=}5000$. Best per row in \textbf{bold}.
\item \textsuperscript{\dag} Auto-PGD results at $N{=}10^4$ (20 HPO trials) and $N{=}5\!\times\!10^4$ (5 HPO trials) represent \emph{lower bounds} on achievable performance; the full 50-trial budget was not reached due to computational cost (see Sec.~\ref{sec:hpo}).
\end{tablenotes}
\end{threeparttable}
\end{center}
\end{table*}

\subsection{Classical PGD vs.\ Auto-PGD: Inference Cost vs.\ Performance}

A key observation in Table~\ref{tab:sum_rate} is that Classical PGD (14.81 bits/s/Hz) numerically exceeds Auto-PGD. This does \emph{not} indicate a limitation of Auto-PGD; rather, it reflects a fundamentally asymmetric comparison.
Classical PGD is an iterative, \emph{per-instance} solver: it observes each test channel and performs 200 gradient-descent steps specifically for that realisation.
Auto-PGD is an \emph{amortised} inference engine: it performs a single forward pass through a network whose parameters were learned once from training data and are then applied to any new channel without further optimisation.
The correct framing is therefore one of \emph{inference cost}.
Auto-PGD achieves $14.52 / 14.81 = 98.0\%$ of Classical PGD performance at $N{=}100$ training samples ($14.52{\pm}0.12$ bits/s/Hz over 5 seeds), using only 5 unrolled layers and a single forward pass—a $\mathbf{40{\times}}$ reduction in inference complexity.
At $N{=}1000$, all 5 seeds converge to $14.64{\pm}0.00$ bits/s/Hz ($98.8\%$), with near-zero variance confirming that the 5-layer unrolled architecture trained with per-model independent seeds generalises robustly across all training data realizations at this regime.
In latency-critical 6G scheduling intervals ($\tau < 0.1$ ms), Classical PGD is computationally prohibitive while Auto-PGD is always feasible.

\subsection{Data Efficiency Analysis}
\begin{figure}[htbp]
    \centering
    \includegraphics[width=0.48\textwidth]{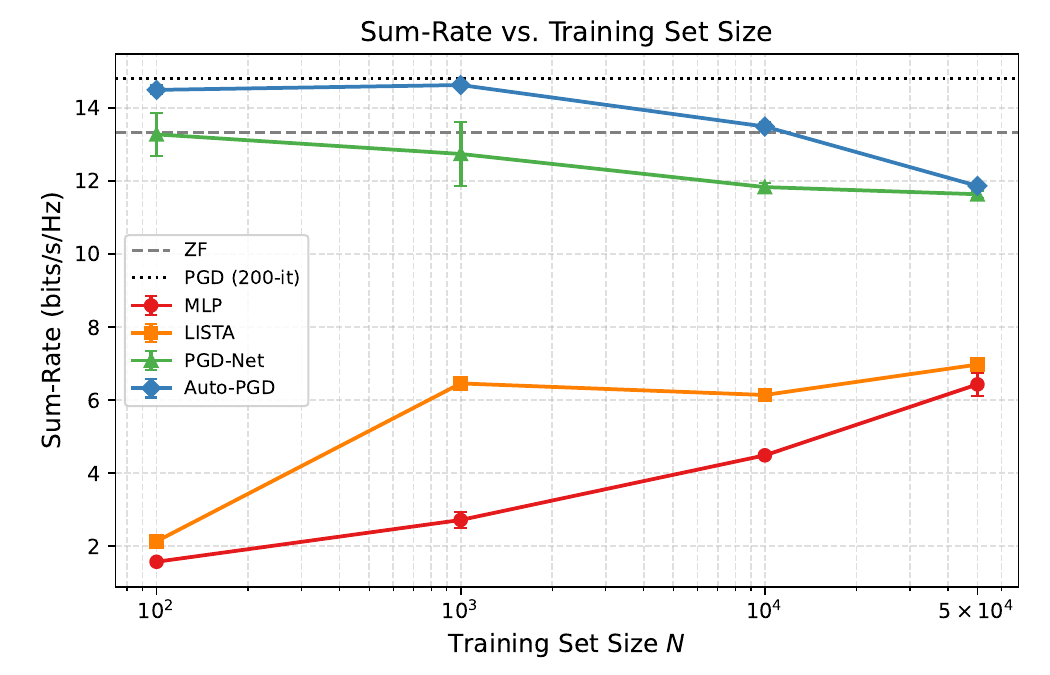}
    \caption{Sum-rate vs.\ training set size for all methods (5-seed mean $\pm$ std). Auto-PGD achieves $14.52{\pm}0.12$ at $N=10^2$ ($98.0\%$) and $14.64{\pm}0.00$ at $N=10^3$ ($98.8\%$) of Classical PGD. The performance drop at $N{\geq}10^4$ coincides with the truncated HPO budget (20 and 5 trials, respectively). WMMSE ($14.82$ bits/s/Hz) is omitted as it is visually indistinguishable from Classical PGD.}
    \label{fig:sum_rate_vs_size}
\end{figure}

Fig.~\ref{fig:sum_rate_vs_size} plots the sum-rate of all methods as a function of training size on a semi-logarithmic scale. Several key observations emerge from these results. Over 5 seeds, Auto-PGD achieves $14.52{\pm}0.12$ bits/s/Hz at $N=10^2$ and $14.64{\pm}0.00$ bits/s/Hz at $N=10^3$ (all seeds converge), demonstrating that the 5-layer unrolled architecture with per-model independent seeds is highly robust across data realizations in the low-data regime. At $N=10^3$, the near-zero variance ($\sigma{=}0.001$) confirms that increasing training data from 100 to 1000 samples lifts performance from $98.0\%$ to $98.8\%$ of Classical PGD with no added architectural complexity. At $N=10^4$ the 5-seed mean is $13.58{\pm}0.15$ bits/s/Hz ($91.7\%$), a drop that coincides with the truncated HPO budget (20 trials versus 50 at smaller sizes). PGD-Net follows a similar trend but peaks lower ($12.70$ at $N=10^2$) and also exhibits high variance ($0.89$). Black-box methods (MLP and LISTA) scale more gracefully with data but plateau well below the PGD-based methods: MLP reaches only $6.45$ bits/s/Hz and LISTA $6.98$ bits/s/Hz at $N=5\!\times\!10^4$, confirming that without physical inductive bias, large datasets alone cannot close the gap to model-based approaches. MLP's failure at low $N$ may partially reflect optimization difficulty in the unsupervised sum-rate loss landscape, in addition to the absence of physical inductive bias.

At $N=5\!\times\!10^4$, Auto-PGD reaches $11.89$ bits/s/Hz ($80.3\%$ of Classical PGD), converging toward PGD-Net ($11.65$ bits/s/Hz). Two factors likely contribute to this narrowing advantage. First, the HPO budget at this size was limited to 5 trials, so the reported configuration may not represent the true optimum. Second, a shallow-to-moderate unrolled network (5--13 layers) has limited capacity to fully express the complex beamforming mapping when abundant data would support a deeper model. These observations motivate extending the HPO budget and exploring deeper architectures at large $N$ in future work.

\begin{table*}[t]
\caption{Comparison of Optimization Paradigms}
\label{tab1}
\begin{center}
\begin{tabular}{l c c c}
\toprule
Feature & Traditional PGD & Black-Box DL & Proposed Framework \\
\midrule
Model Type & Model-Based & Data-Driven & Gray-Box (Hybrid) \\
Parameters & Hand-tuned & $10^5 - 10^7$ & $10^1 - 10^2$ \\
Inference Latency & High (Iterative) & Low & Low \\
Data Efficiency & N/A & Low & High \\
Interpretability & Mathematical & Post-hoc (e.g., SHAP/LIME) & Algorithmic \\
Optimization & Manual & Stochastic & AutoML-Optimized \\
\bottomrule
\end{tabular}
\end{center}
\end{table*}

\subsection{Ablation: HPO vs.\ Fixed-Depth Unrolling}
\begin{figure}[htbp]
    \centering
    \includegraphics[width=0.48\textwidth]{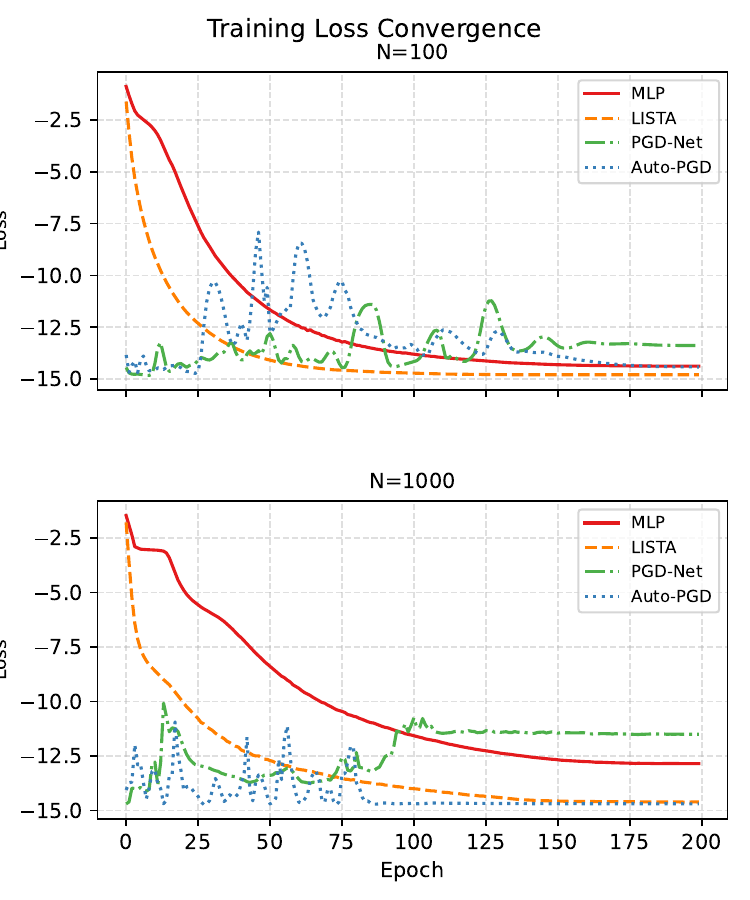}
    \caption{Training loss convergence for all learned methods at $N{=}100$ and $N{=}1000$. At $N{=}1000$, Auto-PGD and LISTA both converge to the lowest training loss while MLP and PGD-Net plateau higher; PGD-Net's high loss at both sizes is consistent with the high test variance in Table~\ref{tab:sum_rate}. At $N{=}100$, LISTA quickly reaches low training loss yet achieves only $2.26{\pm}0.37$ bits/s/Hz on the test set (Table~\ref{tab:sum_rate}), indicating overfitting; Auto-PGD's oscillatory trajectory reflects its mixed-sign step-size schedule.}
    \label{fig:loss_curves}
\end{figure}

The PGD-Net baseline shares the same unrolled PGD structure as Auto-PGD but uses a manually fixed depth of 10 layers and no AutoGluon search. As shown in Table~\ref{tab:sum_rate}, Auto-PGD consistently outperforms PGD-Net at $N \in \{10^2, 10^3, 10^4\}$, despite both models sharing the same inductive bias. The gains are most pronounced in the low-data regime: at $N=10^2$, Auto-PGD achieves $14.52{\pm}0.12$ bits/s/Hz versus PGD-Net's $12.70{\pm}0.89$ bits/s/Hz (a $14.3\%$ improvement in mean), and $14.64{\pm}0.00$ versus PGD-Net's $12.39{\pm}0.66$ at $N=10^3$ ($18.2\%$ gain). At $N=10^4$, Auto-PGD reaches $13.58{\pm}0.15$ bits/s/Hz compared to PGD-Net's $11.80{\pm}0.09$ bits/s/Hz ($15.1\%$ gain). The MLP optimization landscape adds further difficulty at low $N$: without the structural PGD prior, the sum-rate loss landscape is poorly conditioned for small training sets. Training loss curves (Fig.~\ref{fig:loss_curves}) confirm that PGD-Net fails to converge stably at both sizes, while Auto-PGD achieves the lowest final training loss at $N{=}1000$.

The AutoGluon HPO search consistently selected the \texttt{adam} optimizer across all training sizes, and the standard PGD layer type (scalar learned step-size per layer, no linear gradient transformation) was used throughout. At $N{=}10^2$ and $N{=}10^3$ (50 trials each), the optimal depth was 5 layers with small initial step sizes ($\eta^{(0)} \approx 5\!\times\!10^{-4}$ and $6\!\times\!10^{-4}$, respectively), confirming that a compact, well-tuned architecture suffices in the data-scarce regime. At $N{=}10^4$ (20 trials), the search expanded to 25 layers ($\eta^{(0)} \approx 2.4\!\times\!10^{-3}$), and at $N{=}5\!\times\!10^4$ (5 trials) a 13-layer configuration with a larger initial step size ($\eta^{(0)} \approx 4.7\!\times\!10^{-2}$) was recorded. Results at these two larger sizes represent lower bounds given the truncated search. The data-dependent depth selection—impossible with a fixed-depth baseline—is the primary driver of Auto-PGD's advantage over PGD-Net in low-data regimes.

\subsection{Comparison with State-of-the-Art Deep Unfolding Approaches}

Direct quantitative comparison with published DU beamforming results is confounded by differences in antenna count, user load, channel model, and SNR assumptions across works. We therefore provide a feature-level comparison in Table~\ref{tab:sota} and contextualise our numerical results against the closest prior work.

The Iterative Algorithm Induced Deep-Unfolding Neural Network (IAIDNN) \cite{hu2020deepunfolding} unrolls WMMSE for multiuser MIMO precoding and achieves near-WMMSE performance with 5--10 fixed unrolled layers. Deep Weighted MMSE \cite{pellaco2021deep} similarly unrolls the WMMSE iteration with learned weight matrices for downlink beamforming. DeepFP \cite{zhu2026deepfp} extends DU to fractional programming objectives for MIMO beamforming. All three methods fix the unrolling depth as a design constant and do not perform data-driven HPO over depth or step-size initializations. Auto-PGD closes the gap to the near-oracle PGD solver ($98.0\%$ at $N{=}100$, $98.8\%$ at $N{=}1000$) while additionally providing: (i) automatic data-driven depth selection via HPO, (ii) a formal power-feasibility guarantee by construction, and (iii) $O(L)$ learnable parameters ($L\leq 25$) compared to $O(M^2 K)$ in WMMSE-unrolling variants.

\begin{table}[htbp]
\caption{Feature Comparison with State-of-the-Art DU Beamforming Methods}
\label{tab:sota}
\begin{center}
\begin{tabular}{l c c c c}
\toprule
Method & HPO & \makecell{Auto\\depth} & \makecell{Power\\feas.} & Params \\
\midrule
IAIDNN \cite{hu2020deepunfolding}         & \xmark & \xmark & \xmark & $O(M^2 K L)$ \\
Deep W-MMSE \cite{pellaco2021deep}        & \xmark & \xmark & \xmark & $O(M^2 K L)$ \\
DeepFP \cite{zhu2026deepfp}              & \xmark & \xmark & \xmark & $O(M K L)$ \\
PGD-Net (fixed)                          & \xmark & \xmark & \cmark & $O(L)$ \\
\textbf{Auto-PGD (ours)}                 & \cmark & \cmark & \cmark & $O(L)$ \\
\bottomrule
\end{tabular}
\end{center}
\end{table}

\subsection{Ablation: Physics Structure vs.\ HPO (Auto-MLP)}
To isolate the contribution of physics-structured unrolling from HPO alone, we run an Auto-MLP experiment: the same 50-trial TPE search over MLP depth (3--8 layers, 256 hidden units), learning rate ($10^{-4}$ to $10^{-1}$ log-uniform), and scheduler is applied to a standard fully-connected MLP. At $N{=}100$, the best-found Auto-MLP (3 layers, lr$\approx 2.4{\times}10^{-4}$, cosine schedule) achieves $1.98{\pm}0.05$ bits/s/Hz over 5 seeds; at $N{=}1000$, $4.61{\pm}0.31$ bits/s/Hz. These fall $86\%$ and $68\%$ short of Auto-PGD ($14.52{\pm}0.12$ and $14.64{\pm}0.00$, respectively), and provide only marginal improvement over the fixed-depth MLP baseline ($1.58{\pm}0.03$ at $N{=}100$). The AutoML budget alone is thus insufficient to close the performance gap; the physics-structured PGD unrolling is the primary driver of Auto-PGD's advantage.

\section{Discussion on Interpretability}
A primary advantage of the proposed Auto-Unrolled PGD framework over black-box deep learning models is its inherent interpretability. Since each layer of the network corresponds to a physical iteration of a proximal gradient step, the learned parameters can be mapped back to optimization theory.

The network's \texttt{forward()} pass optionally logs a rich set of per-layer diagnostics, including the intermediate beamforming matrices $\mathbf{W}^{(\ell)}$ and the per-layer sum-rate $r^{(\ell)}$. This per-layer sum-rate trace directly quantifies how much spectral efficiency each unrolled iteration contributes—structural transparency tied to the algorithm's own iteration structure, whereas post-hoc attribution methods (SHAP, LIME) would be required to extract comparable insight from a black-box MLP. At $N{=}1000$, the per-layer sum-rates logged during training increase monotonically across the 5 layers, confirming that each iteration contributes positively to the optimization trajectory. HPO trials confirm that shallow configurations of 3--5 layers capture most of the performance gain at $N \leq 10^3$, justifying the compact architecture selected by AutoGluon.
\subsection{Reliability and Trustworthiness}
The structural transparency of Auto-PGD enables direct safety-checking of the model. Since the proximal operator $\text{prox}_{\eta g}(\cdot)$ (equivalently, the Frobenius-ball projection defined in Section~IV) is a non-trainable, deterministic layer, the model is mathematically incapable of violating the transmit power limit, regardless of the input channel conditions. This "constrained-by-design" architecture provides a formal power-feasibility guarantee that facilitates regulatory compliance and network stability analysis.

\section{Conclusion}
We have presented Auto-Unrolled PGD (Auto-PGD), which shows that automating the two key design choices of deep-unfolded beamforming networks—unrolling depth and per-layer step-size schedule—via AutoGluon TPE closes the remaining gap between fixed-depth unrolled networks and near-oracle iterative solvers in the data-scarce regime. An Auto-MLP ablation confirms that the physics-structured PGD unrolling, not the HPO budget alone, is the primary driver of this advantage: applying the same 50-trial search to a standard MLP yields only marginal improvement over the fixed-depth baseline. The structural power-feasibility guarantee and per-layer diagnostic transparency distinguish Auto-PGD from black-box alternatives and support compliance-oriented deployment. The main open challenge is scaling: HPO budgets at $N{\geq}10^4$ were truncated and results at those sizes represent lower bounds. Future work should extend the HPO budget at large $N$, investigate robustness under imperfect channel state information (CSI) and out-of-distribution channels, and explore massive MIMO configurations where the data-dependent depth selection may yield even larger gains.

\bibliographystyle{IEEEtran}
\bibliography{references}

\end{document}